\documentclass{article}
\usepackage{wrapfig,lipsum,booktabs}
\usepackage{times}
\usepackage{epsfig}
\usepackage{graphicx}
\usepackage{amsmath}
\usepackage{amssymb}
\usepackage{subcaption}
\usepackage{boldline,multirow}
\usepackage{booktabs} 
\usepackage{comment}

\usepackage[pagebackref=true,breaklinks=true,letterpaper=true,colorlinks,bookmarks=false]{hyperref}
\usepackage[nonatbib,final]{neurips_2019}
\usepackage{amsmath}

\newcommand{\norm}[1]{\left\lVert#1\right\rVert}

\begin{document}
	
	\title{Localizing Occluders with \\Compositional Convolutional Networks}
	
	\author{
		Adam Kortylewski, 
		Qing Liu, 
		Huiyu Wang, 
		Zhishuai Zhang,
		Alan Yuille\\
		Johns Hopkins University\\
		{\tt\small \{akortyl1,qingliu,hwang157,zzhang99,ayuille1\}@jhu.edu}
	}
	\maketitle
\begin{abstract}
	Compositional convolutional networks are generative compositional models of neural network features, that achieve state of the art results when classifying partially occluded objects \cite{kortylewski2019compositional}, even when they have not been exposed to occluded objects during training. 
	In this work, we study the performance of CompositionalNets at localizing occluders in images. We show that the original model \cite{kortylewski2019compositional} is not able to localize occluders well. We propose to overcome this limitation by modeling the feature activations as a mixture of von-Mises-Fisher distributions, which also allows for an end-to-end training of CompositionalNets. Our experimental results demonstrate that the proposed extensions increase the model's performance at localizing occluders as well as at classifying partially occluded objects.
\end{abstract}

	\section{Introduction}
In natural images, objects are surrounded and partially occluded by other objects. 
Current deep models are significantly less robust to partial occlusion compared to Humans \cite{hongru,kortylewski2019compositional}.
Kortylewski et al. recently proposed the compositional convolutional network (CompositionalNet) \cite{kortylewski2019compositional}, a generative compositional model of neural feature activations that can classify partially occluded objects with exceptional performance, even when it has not been exposed to occluded objects during training.

While robustness to partial occlusion is a desirable property of vision systems, they should also be able to localize which parts of the object are occluded.
The ability to localize occluders in an image is important because it improves explainability of the classification process and enables future research on parsing scenes with mutually occluding objects. Preliminary results \cite{kortylewski2019compositional} indicate the potential of CompositionalNets at localizing occluders, but this remains to be confirmed quantitatively.

In this work, we study the ability of CompositionalNets at localizing occluders in an image.
Our experiments show that CompositionalNets as proposed in \cite{kortylewski2019compositional} are not able to localize occluders well (see experiments in Section \ref{sec:exp}), despite being able to classify partially occluded images robustly.
We find that the reason for this is a simplified assumption about the distribution of neural feature activations in CompositionalNets (feature activations are binarized and their distribution was modeled using a Bernoulli distribution \cite{kortylewski2019compositional}).
We propose to overcome this limitation by modeling the real-valued feature activations as a mixture of von Mises Fisher distributions, which also enables an end-to-end training of all parameters of the neural network. We demonstrate that these extensions enhance the ability of CompositionalNets at localizing occluders significantly while also increasing their exceptional performance at classifying partially occluded objects.

Our work demonstrates that deep networks with a generative compositional architecture have strong generalization abilities that allow them to classify partially occluded objects robustly and to localize occluders in images, even when they have not been exposed to partially occluded objects during training.

\section{Compositional Convolutional Networks}
In this section, we first review CompositonalNets as introduced in \cite{kortylewski2019compositional} (Section \ref{sec:comp}) and then propose an extension that enables them to better localize occluders in an image and to be trained in an end-to-end manner (Section \ref{sec:vmf}).

\begin{figure}
	\begin{subfigure}{0.5\linewidth}
		\includegraphics[width=\linewidth]{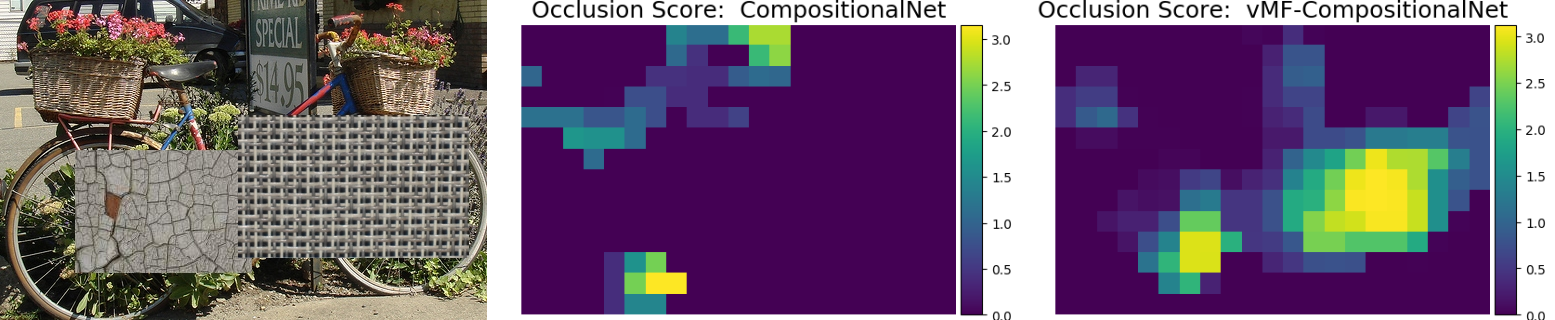}
	\end{subfigure}
	\begin{subfigure}{0.48\linewidth}
		\includegraphics[width=\linewidth]{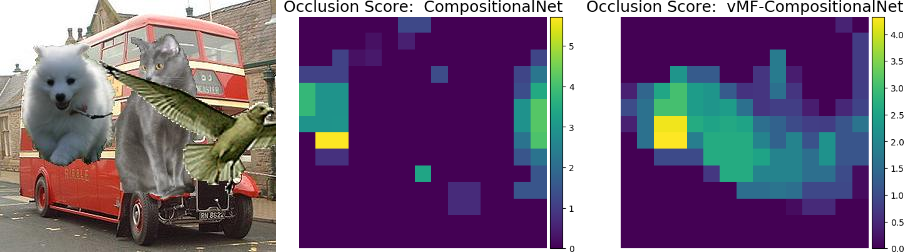}		
	\end{subfigure}
	\caption{ 
		Occluder localization with CompositionalNets. Each result consists of three images: The input image (left) and occlusion scores based on feature activations in the pool4 layer of standard CompositionalNets \cite{kortylewski2019compositional} (middle) and our proposed extension respectively (right).
		We can clearly observe that our proposed model can better localize occluders in the images.
	} 
	\label{fig:intro}
\end{figure}	

\subsection{CompositionalNets with Dictionary Encodings}
\label{sec:comp}
We define a feature map $F^l$ as the output of a layer $l$ in a CNN. 
A feature vector $f^l_p \in \mathbb{R}^C$ is the vector of features in $F^l$ at position $p$, where $p$ is defined on the 2D lattice of the feature map and $C$ is the number of channels in the layer. 
Note that the spatial information from the image is preserved in the feature maps, thus a position $p$ on $F^l$ corresponds to a patch in the image. 
We omit the subscript $l$ in the remainder of this section because the layer from which the features are extracted is fixed in our model (e.g. $l=4$ for the layer $conv_4$).

\textbf{A generative model of binary dictionary encodings.} The authors in \cite{kortylewski2019compositional} proposed to encode the feature maps $F$ with a dictionary $D=\{d_1,\dots,d_K\}$. The dictionary is learned by clustering the vectors of the feature maps from all training images $\{F^n | n=1,\dots,N\}$. The feature vectors $f_p$ are encoded with a sparse binary vector $b_p$ by detecting the nearest neighbors of $f_p$ in the learned part dictionary $D$ using the cosine distance $g(\cdot|\cdot)$. 
Hence, the element $b_{p,k} = 1$ if $g(f_p,d_k)>\delta$.
Intuitively, $b_p$ encodes which parts of the dictionary $D$ are detected at position $p$ in the feature map $F$. A generative model of the binary activation matrix $B$ is defined as Bernoulli distribution:
\begin{align}
\label{eq:fg}
p(B|\mathcal{A}_y) =  \prod_{p} p(b_p|\alpha_{p,y})
= \prod_{p,k} \alpha_{p,k,y}^{b_{p,k}} (1-\alpha_{p,k,y})^{1-b_{p,k}}.
\end{align}		
Where $\alpha_{p,k,y}$ is the probability that the part $d_k$ is active at position $p$ for the object class $y$, and thus $b_{p,k}=1$.

\textbf{Mixture of compositional models.} 
Because of the independence between parts in Eq. \ref{eq:fg}, the model assumes that the spatial distribution of parts in $B$ is approximately the same.
This assumption does not hold for 3D objects, because e.g. by changing the 3D pose of an object the relative spatial distribution of parts also changes strongly. This issue can be resolved using mixtures of compositional models: 
\begin{align}
p(B|\mathcal{A}_y,\mathcal{V}) = \prod_m p(B|\mathcal{A}^m_y)^{\nu_m},
 \sum_m \nu_m = 1,\hspace{.1cm} \nu_m \in\{0,1\}.
\end{align}
The intuition is that each mixture component $m$ will represent images of an object that have approximately the same spatial part distribution (i.e. similar viewpoint and 3D structure).
The parameters of the individual mixtures $A^m_y$ as well as the mixture assignment variables $\mathcal{V}$ can be learned using maximum likelihood estimation while alternating between estimating $A^m_y$ and $\mathcal{V}$. 

\textbf{Occlusion modeling.}
Partial occlusion of an object will change the part activation patterns in $B$ such that parts may be missing and other parts might be active at a previously unseen location. 
The intuition behind an occlusion model is that at each position $p$ in the image either the object model $\mathcal{A}_y$ or a background model $\beta$ is active:
\begin{align}
\label{eq:occ}
	p(B|\Gamma) = \prod_{p} p(b_p|FG)^{z_p} p(b_p|BG)^{1-z_p},\hspace{0.1cm}z_p \in \{0,1\}, \hspace{0.1cm}\Gamma=\{\mathcal{A}_y;\beta;\mathcal{Z}\}\\
	p(b_p|FG) = p(b_p|\alpha_{p,y}) p(z_p),\hspace{0.1cm}p(b_p|BG) = p(b_p|\beta) (1-p(z_p)).
\end{align}

The binary variable $z_p$ indicates if the object is visible at position $p$. The occlusion prior $p(z_p)$ can be learned or alternatively be set manually (see Section \ref{sec:exp}). The background model is defined as:
$	p(b|\beta) = \prod_{k} \beta_{k}^{b_{k}} (1-\beta_{k})^{1-b_{k}}$.
Here the background model is assumed to be independent of the position in the image and thus has no spatial structure. 
The background model can be estimated by $\beta=\frac{1}{J}\sum_{j=1}^J b_j$, where $J$ part detection vectors $b_j$ are randomly sampled on a set of background images that do not contain one of the objects of interest. 

\subsection{Fully Generative CompositionalNets}
\label{sec:vmf}
The model as presented in the previous section can recognize partially occluded objects with high-performance \cite{kortylewski2019compositional}, however, it is not able to discriminate well between the occluder and the object (see experiments in Section \ref{sec:exp}). 
One reason is that instead of modeling the distribution of the real-valued features $p(F|y)$, the authors in \cite{kortylewski2019compositional} binarize the features with a heuristic threshold and just model the distribution of binary activations $p(B|y)$ (Equation \ref{eq:fg}). The binarization has two disadvantages: 1) Some information that is useful to discriminate between the object and the occluder is lost. 2) The thresholding operation is not differentiable and therefore prevents an end-to-end optimization of the model. We propose to replace the Bernoulli distribution over binary features (see Equation \ref{eq:fg}) with a mixture of von Mises Fisher (vMF) distributions:
\begin{align}
\label{eq:vmf}
p(F|\Theta_y) =  \prod_{p} p(f_p|\mathcal{A}_{p,y},\theta) = \prod_{p} \sum_k \alpha_{p,k,y} p(f_p|S_k,\mu_k),
\end{align}
where $\Theta_y= \{\mathcal{A}_{0,y},\dots,\mathcal{A}_{\mathcal{P},y},\theta\}$ are the model parameters at every position $p \in \mathcal{P}$ on the lattice of the feature map $F$, $\mathcal{A}_{p,y} = \{\alpha_{p,0,y},\dots,\alpha_{p,K,y}|\sum_{k=0}^K \alpha_{p,k,y} = 1\}$ are the mixture coefficients, $K$ is the number of mixture components, $\theta = \{\theta_k = \{S_k,\mu_k \} | k=1,\dots,K \}$ are the parameters of the vMF mixture distributions:
\begin{equation}
p(f_p|S_k,\mu_k) = \frac{e^{S_k \mu_k^T f_p}}{Z(S_k)}, \norm{f_p} = 1, \norm{\mu_k}= 1, 
\end{equation}
and $Z(S_k)$ is the normalization constant. Accordingly, we define the background model as
$p(f|\beta) = \sum_k \beta_{k} p(f|S_k,\mu_k)$. The parameters of the vMF mixture model $\Theta_y$ and the background model $\beta$ can be learned with maximum likelihood estimation, as discussed in \cite{kortylewski2019compositional}. Compared to the model in \cite{kortylewski2019compositional}, our fully generative model avoids the binarization of features and therefore can be fine-tuned in an end-to-end manner.

\begin{table*}
	\small
	\centering
	\tabcolsep=0.11cm
	\begin{tabular}{lV{2.5}cV{2.5}c|c|c|cV{2.5}c|c|c|cV{2.5}c|c|c|cV{2.5}c}
		\multicolumn{15}{c}{\textbf{Classification under Occlusion}} \\
		\hline
		Occ. Area 					         & \textbf{0\%} & \multicolumn{4}{cV{2.5}}{\textbf{Level-1: 20-40\%}} & \multicolumn{4}{cV{2.5}}{\textbf{Level-2: 40-60\%}} & \multicolumn{4}{cV{2.5}}{\textbf{Level-3: 60-80\%}} & Mean \\
		\hline
		Occ. Type 	& - & w & n & t & o & w & n & t& o & w & n & t & o &-\\    		
		\hline  
		VGG 		& 99.2 &97.9&97.9&97.6&90.3&91.6&90.5&89.7&68.8&54.7&52.3&48.1&47.5&78.9\\		
		\hline
		CompMixOcc-Dict &92.1&92.7&92.3&91.7&92.3&87.4&89.5&88.7&90.6&70.2&80.3&76.9&87.1&87.1\\
		\hline	
		CompMixOcc-Full
		&95.9&95.8&95.2&94.9&94.9&95.0&93.3&92.9&92.3&86.8&83.8&80.9&88.1&91.5\\	
		\hline					
		CompNet-Dict
		&98.3&96.8&95.9&96.2&94.4&91.2&91.8&91.3&91.4&71.6&80.7&77.3&87.2&89.5		
		\\
		\hline
		CompNet-Full
		&98.6&97.9&97.5&97.3&96.1&95.9&94.5&94.1&92.4&86.8&84.0&80.9&87.7&92.6\\
		\hline		
		\hline		
		Human & 100.0& \multicolumn{4}{cV{2.5}}{100.0}& \multicolumn{4}{cV{2.5}}{100.0}  & \multicolumn{4}{cV{2.5}}{98.3}& 99.5
		\vspace{.2cm}
	\end{tabular}
	\caption{Object classification under occlusion. The proposed fully generative model outperforms the dictionary-based model proposed in \cite{kortylewski2019compositional}, while it also performs much better at occlusion localization (see Figure \ref{fig:occ-det}).}
	\label{tab:occ}	
\end{table*}

\section{Experiments}
\label{sec:exp}
\begin{figure*}
	\begin{subfigure}{0.49\linewidth}
		\centering
		\includegraphics[width=\linewidth]{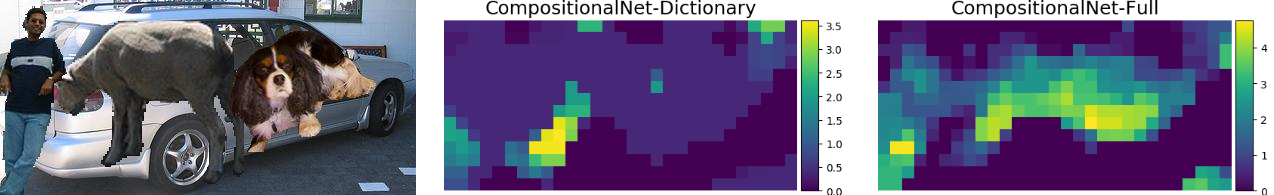}
	\end{subfigure}
	\begin{subfigure}{0.49\linewidth}
		\centering
		\includegraphics[width=\linewidth]{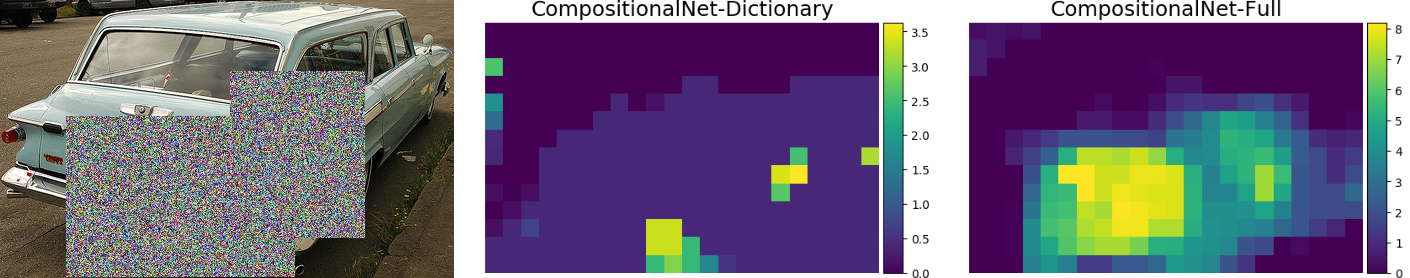}
	\end{subfigure}\\\\\\
	\begin{subfigure}{0.49\linewidth}
		\centering
		\includegraphics[width=\linewidth]{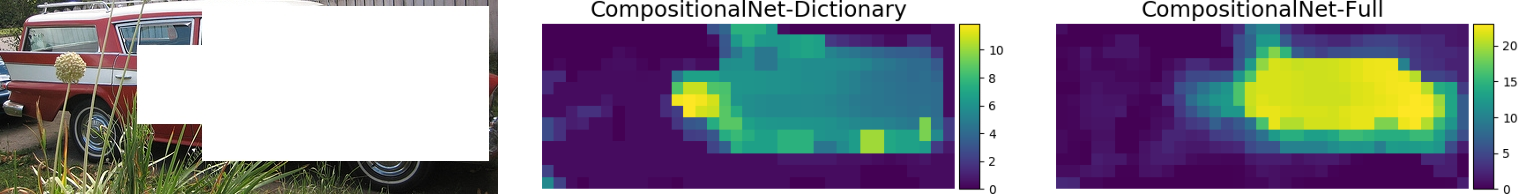}
	\end{subfigure}
	\begin{subfigure}{0.49\linewidth}
		\centering
		\includegraphics[width=\linewidth]{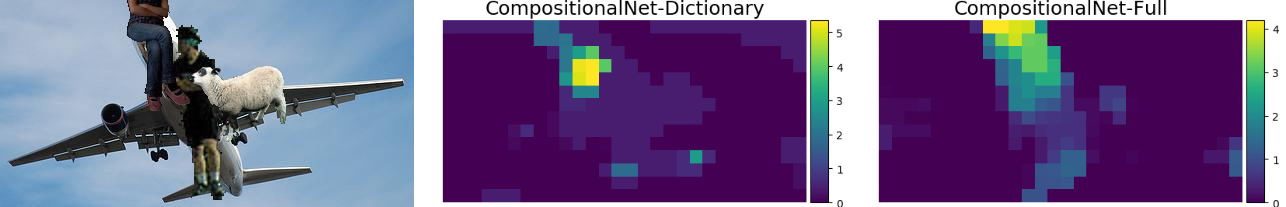}
	\end{subfigure}\\\\\\
	\begin{subfigure}{0.49\linewidth}
		\centering
		\includegraphics[width=\linewidth]{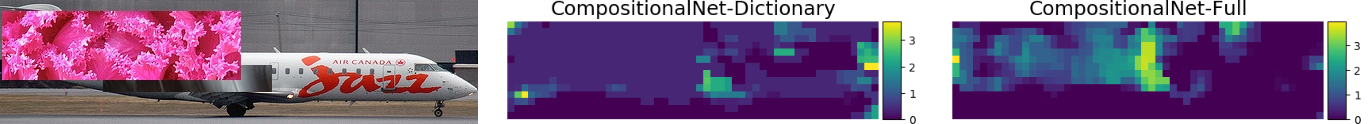}
	\end{subfigure}
	\begin{subfigure}{0.49\linewidth}
		\centering
		\includegraphics[width=\linewidth]{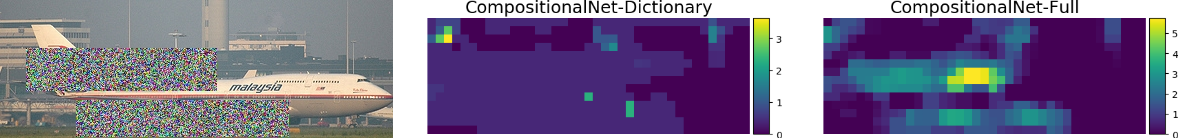}
	\end{subfigure}
	\caption{Visualization of occlusion localization results (not cherry picked).
	Each result consists of three images: The input image, the occlusion scores of a dictionary-based CompositionalNet and our proposed fully generative CompositionalNet. Note how our model can localize occluders with higher certainty across objects and occluder types.}
	\label{fig:occ-quali}
\end{figure*}

In this section, we compare our proposed fully generative CompositionalNet with the dictionary-based model as described in Section \ref{sec:comp} at object classification and occluder localization. The experiments are performed on the OccludedVehicles dataset proposed in \cite{wang2015unsupervised} and extended in \cite{kortylewski2019compositional}. The dataset consists of vehicles from the from the PASCAL3D+ dataset \cite{xiang2014beyond} that were synthetically occluded by four different types of occluders (see Figure \ref{fig:occ-quali}): 
\textit{real objects} and patches with \textit{constant white color}, \textit{random noise} and \textit{textures}. At training time all models are trained on non-occluded images, while at test time the models are exposed to images with different amount of partial occlusion. The CompositionalNets are trained from the feature activations of the pool4-layer of a VGG model that was pretrained on ImageNet \cite{deng2009imagenet}. Our training setup is chosen as specified in \cite{kortylewski2019compositional}.

\begin{figure*}
	\begin{subfigure}{0.33\linewidth}
		\centering
		\includegraphics[height=4cm]{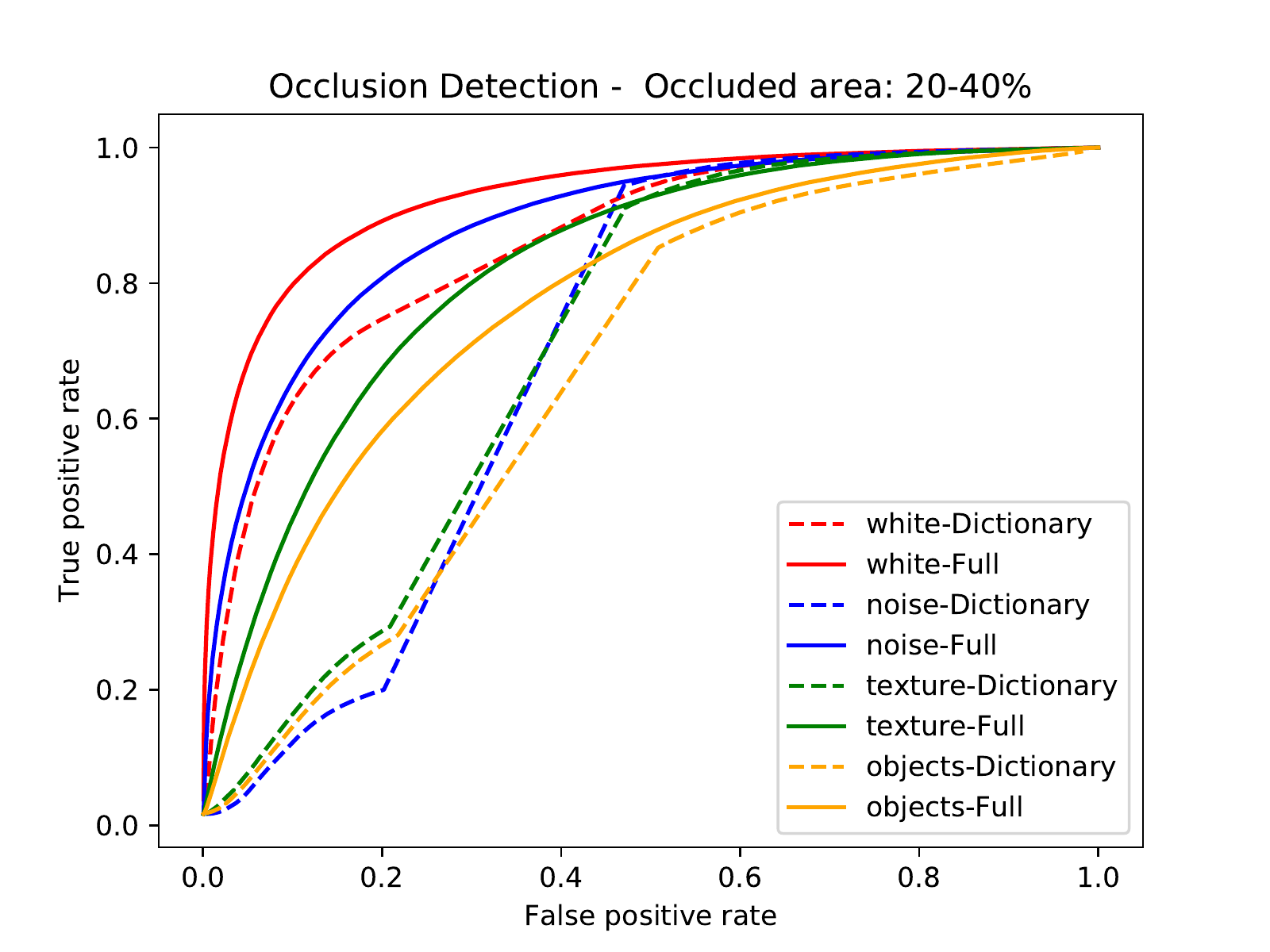}
		\vspace{-.25cm}		
		\caption{}
	\end{subfigure}%
	\begin{subfigure}{0.33\linewidth}
		\centering
		\includegraphics[height=4cm]{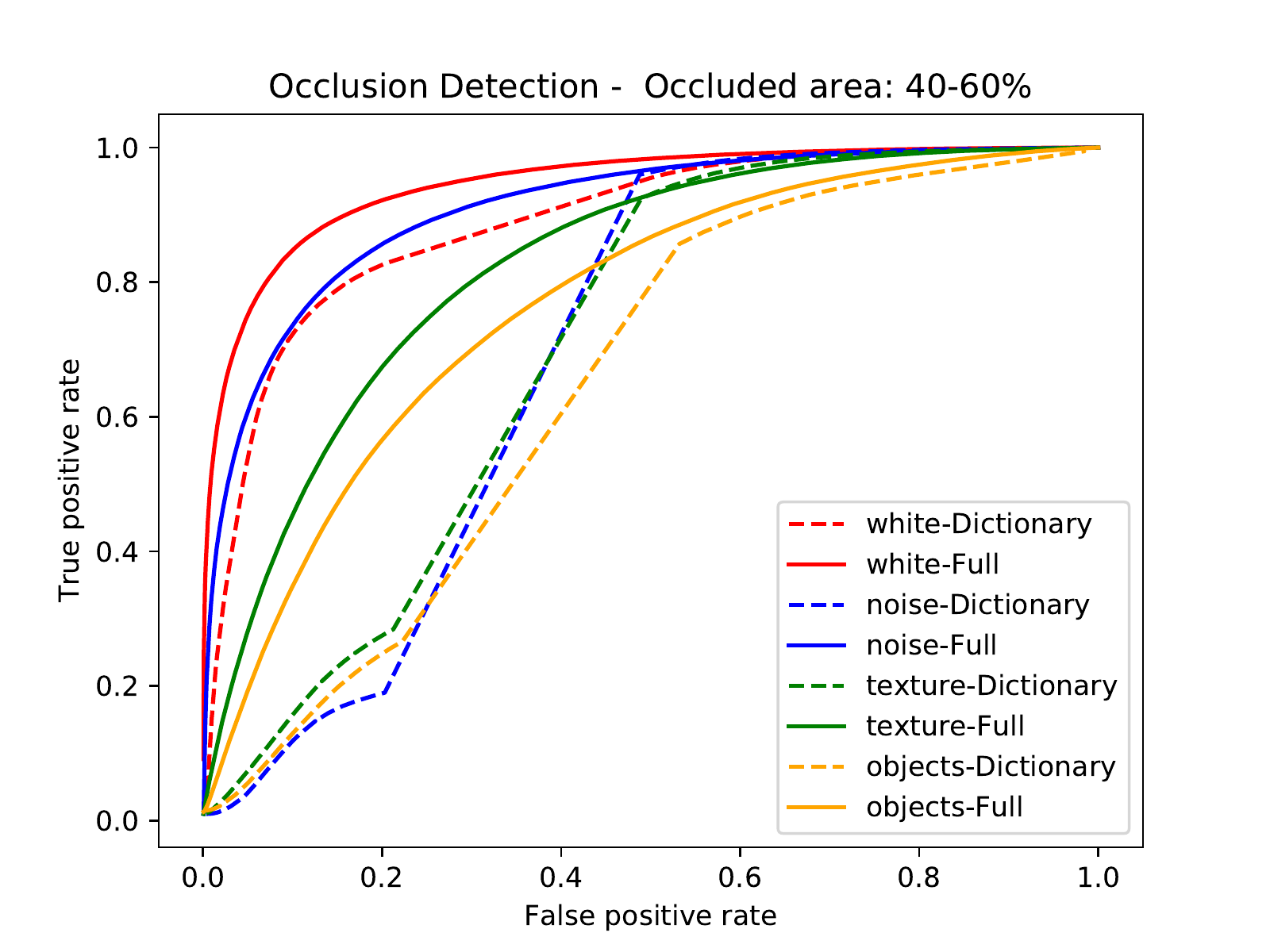}
		\vspace{-.25cm}		
		\caption{}		
	\end{subfigure}%
	\begin{subfigure}{0.33\linewidth}
		\centering
		\includegraphics[height=4cm]{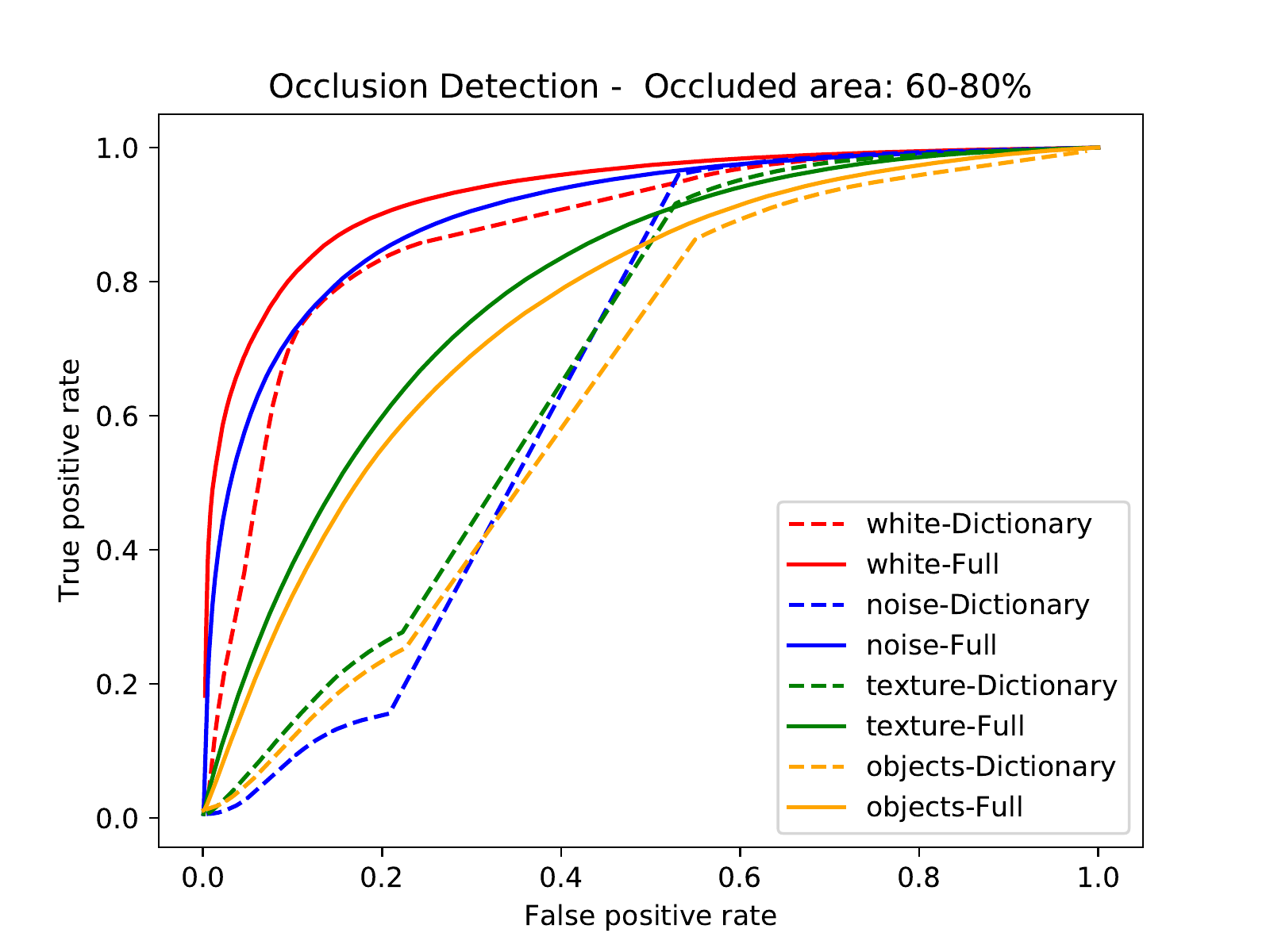}
		\vspace{-.25cm}		
		\caption{}		
	\end{subfigure}
	\vspace{-.25cm}
	\caption{Occluder localization with dictionary-based CompositionalNets and our proposed fully generative CompositionalNet for different levels of partial occlusion: (a) 20-40\%, (b) 40-60\% and (c) 60-80\% of the object is occluded. Our model significantly outperforms dictionary-based CompositionalNets at localizing occluders.}
	\label{fig:occ-det}
	\vspace{-.25cm}
\end{figure*}

\textbf{Localizing occluders with CompositionalNets.}
Figure \ref{fig:occ-quali} illustrates occlusion scores at different positions $p$ of the corresponding feature map $F$. 
We compute the occlusion score as the log-likelihood ratio $\log\frac{p(f_p|BG)}{p(f_p|FG)}$ of the background and foreground model. 
Note that we visualize only positive occlusion scores to highlight the localization of the occluder. 
We can observe that the fully generative model can localize occluders significantly better than in the dictionary-based model for all types of occluders. 
Figure \ref{fig:occ-det} shows the ROC curves of both models when using the occlusion score for classifying each pixel as being occluder or not. 
The dictionary-based model (dotted lines) performs poorly for any type of occluder except the plain white ones. 
In contrast, our fully generative model significantly improves the quality of the occluder classification for all types of occluders. 
The classification results in Table \ref{tab:occ} show that the proposed model also outperforms dictionary-based CompositionalNets at classifying partially occluded objects.


	\section{Conclusion}
We considered the problem of classifying partially occluded objects and localizing the occluders when partially occluded objects are not represented in the training data. 
Our experiments show that dictionary-based CompositionalNets are not able to localize occluders well, although they can classify partially occluded objects robustly. 
We proposed an extension to CompositionalNets that enables them to accurately localize occluders in images while also improving the performance at classifying partially occluded objects. 
Our work shows that neural networks are capable to generalize well beyond the training data in terms of partial occlusion when their architecture is compositional and they are trained to be fully generative in terms of their high-level features.

	{\small
		\bibliographystyle{plain}
		\bibliography{egbib}
	}
\end{document}